\documentclass[letterpaper, conference]{IEEEtran}
\IEEEoverridecommandlockouts
\usepackage{cite}
\usepackage{amsmath,amssymb,amsfonts}
\usepackage{algorithmic}
\usepackage{graphicx}
\usepackage{textcomp}
\usepackage{xcolor}
\usepackage{booktabs}
\usepackage{multirow}
\usepackage{seqsplit}

\usepackage{url}
\usepackage{graphicx}
\def\BibTeX{{\rm B\kern-.05em{\sc i\kern-.025em b}\kern-.08em
    T\kern-.1667em\lower.7ex\hbox{E}\kern-.125emX}}
\usepackage{hyperref}
\hypersetup{
    colorlinks=true,
    allcolors=blue,
    pdftitle={Anomaly Detection and Localization for The Pantograph-Catenary System},
    pdfpagemode=FullScreen,
}

\usepackage{pbalance}

\makeatletter
\newcommand{\linebreakand}{%
  \end{@IEEEauthorhalign}
  \hfill\mbox{}\par
  \mbox{}\hfill\begin{@IEEEauthorhalign}
}
\makeatother
    
\begin{document}

\title{\vspace{0.25in}Anomaly Detection and Localization for the Pantograph-Catenary System
\thanks{This work was supported by the Swiss SERI under contracts no. 23.00321 (Academics4Rail) and 24.00528 (PhDs EU-Rail); those projects have been selected within the EU’s Horizon Europe program under GAs no. 101121842 and 101175856. Views and opinions are those of the authors only and do not necessarily reflect those of the funding agencies, which cannot be held responsible for them.}
}

\author{%
\IEEEauthorblockN{Francesco Vitale}
\IEEEauthorblockA{\textit{DIETI} \\
\textit{University of Naples Federico II}\\
Naples, Italy \\
francesco.vitale@unina.it}
\and
\IEEEauthorblockN{Hangli Ge}
\IEEEauthorblockA{\textit{Interfaculty Initiative in} \\
\textit{Information Studies} \\
\textit{The University of Tokyo}\\
Tokyo, Japan \\
hanglige@g.ecc.u-tokyo.ac.jp}
\and 
\IEEEauthorblockN{Francesco Flammini}
\IEEEauthorblockA{\textit{DIMAI} \\
\textit{University of Florence}\\
Florence, Italy \\
francesco.flammini@unifi.it}
\IEEEauthorblockA{\textit{and IDSIA USI-SUPSI} \\
\textit{University of Applied Sciences and Arts} \\
\textit{of Southern Switzerland}\\
Lugano, Switzerland \\
francesco.flammini@supsi.ch}
}
\maketitle

\begin{abstract}
Monitoring the Pantograph-Catenary System (PCS) provides insight into the health conditions of the pantograph and the railway infrastructure. Recent industrial solutions trace the pantograph's contact wire height and stagger (PCS height/stagger) using video monitoring through convolutional neural networks. However, these solutions do not account for the train route's geographic location. Therefore, in this paper we propose a novel framework for 1) localization of the PCS height/stagger by alignment with the nominal GPS coordinates of the reference route, and 2) collective anomaly detection to evaluate the health conditions of the PCS. We apply and assess the localization and detection performance of the methodology to a case-study based on a real-world industrial dataset provided by a railway transportation company, which includes the PCS height/stagger of several train journeys across Italian railway routes.
\end{abstract}


\section{Introduction}
\label{sec:introduction}
\begin{figure}[!t]
\centering
\includegraphics[width=\columnwidth]{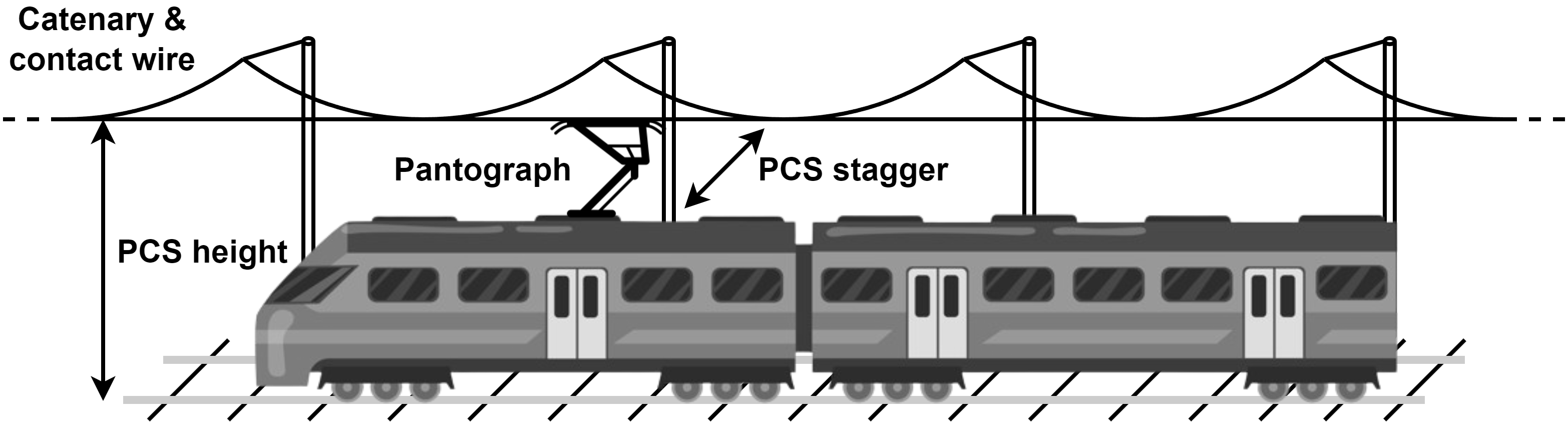}
\caption{The Pantograph-Catenary System.}
\label{PANTOGRAPH_CATENARY_SYSTEM}
\end{figure}

\IEEEPARstart{R}{ailway} infrastructures are critical systems that demand high safety levels, which are achieved by intense verification and validation activities from design time to operation and maintenance
\cite{cimatti2012ertmsfv}. These systems require condition monitoring to assess the health of onboard equipment and infrastructure over time~\cite{chen2024railwaycatenaryconditionmonitoring}. To this aim, several monitoring solutions have been proposed for different systems and devices, including air conditioning and traction systems~\cite{galves2021hvacfaultdetectionrul, wang2022lstmbasedrailtransit}, passenger demand prediction~\cite{ge2024k}.
Modern solutions are increasingly based on machine learning applied to multimedia streams \cite{9795283}, mainly using convolutional neural networks~\cite{Pappaterra2024} or graph neural networks~\cite{hangli2022multi}. In such a context, the work presented in this paper focuses on video monitoring the Pantograph-Catenary System (PCS), which is the system that connects the train's onboard equipment with the railways' overhead catenary for electricity supply through the pantograph's contact wire~\cite{blanco2022pantographstaggermonitoring}. Fig. \ref{PANTOGRAPH_CATENARY_SYSTEM} shows the PCS and its main components, namely the catenary, contact wire and pantograph. The health conditions of the PCS can be assessed by checking the geometric properties of the contact wire, such as the height between the track and catenary and the stagger, i.e., the lateral displacement of the contact wire relative to the centerline of the track~\cite{blanco2022pantographstaggermonitoring}. Anomalous pantograph's contact wire stagger and height (PCS height/stagger) may cause contact wire breakage, pantograph destruction, material damage, fire, and electric shock~\cite{rusuanghel2016faultmodeling}. 

Because camera sensors are versatile and widely used for condition monitoring~\cite{chen2024railwaycatenaryconditionmonitoring}, extensive scientific literature addresses anomaly detection in railway infrastructure using onboard video monitoring, including the PCS. Zhang et al.~\cite{zhang2023locatingralwayscatenaryhangingstring} proposed an improved YOLO Convolutional Neural Network (CNN) to detect catenary defects, specifically discriminating between loosened and broken hanging strings. Focusing on the pantograph, Wei et al.~\cite{wei2020pantographdefectdetectiondl} evaluated several CNNs for classifying wear states (e.g., over-abrasion, partial, and groove-shaped abrasion). Chen et al.~\cite{chen2022pantographad} developed a dedicated CNN to monitor PCS height and stagger, verifying whether these metrics fall within safe operational zones.

Despite advances in video-based PCS monitoring, existing CNN-based solutions track height and stagger without accounting for the train's geographical localization. 
To address this shortcoming, we set forth a framework for collective anomaly detection and localization of the PCS height/stagger with the nominal GPS coordinates of the reference route. We apply the methodology to an industrial case study that collects the PCS height/stagger of several train journeys across railway routes managed by an Italian company.
\section{Reference Framework and Current Results}
\label{sec:framework}
\begin{figure}[!t]
\centering
\includegraphics[width=\columnwidth]{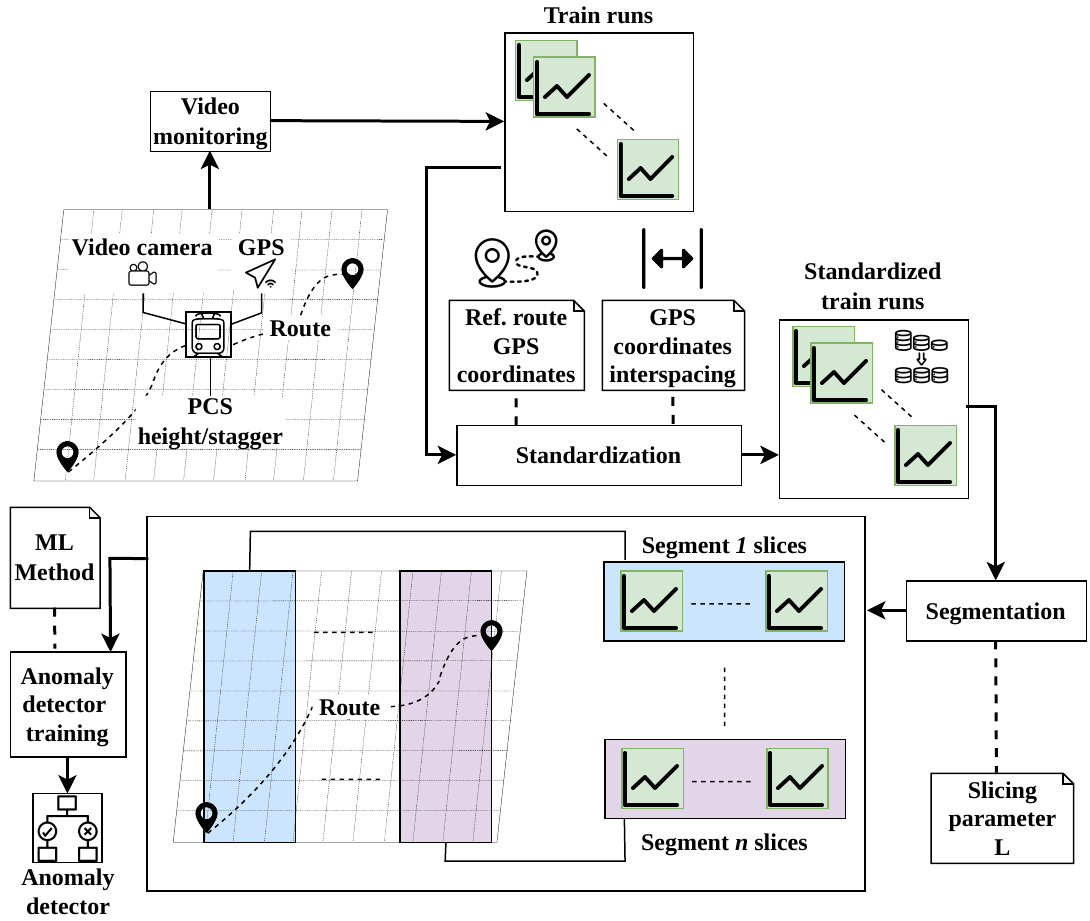}
\caption{The proposed framework for anomaly detection and localization in the PCS height/stagger of train runs.}
\label{METHOD}
\end{figure}

In the proposed framework, depicted in Fig.~\ref{METHOD}, onboard \textbf{video monitoring} serves as the primary data acquisition layer for assessing the health of the PCS. Camera sensors continuously track the contact wire to capture critical geometric metrics, specifically the PCS height/stagger. Crucially, to overcome the limitations of previous solutions that lack spatial awareness, these visual measurements are synchronized in real-time with the train's geographic location using an onboard GPS receiver. This process generates raw logs of individual train runs containing both the physical metrics and their corresponding GPS coordinates.

\begin{figure}[!t]
\centering
\includegraphics[width=\columnwidth]{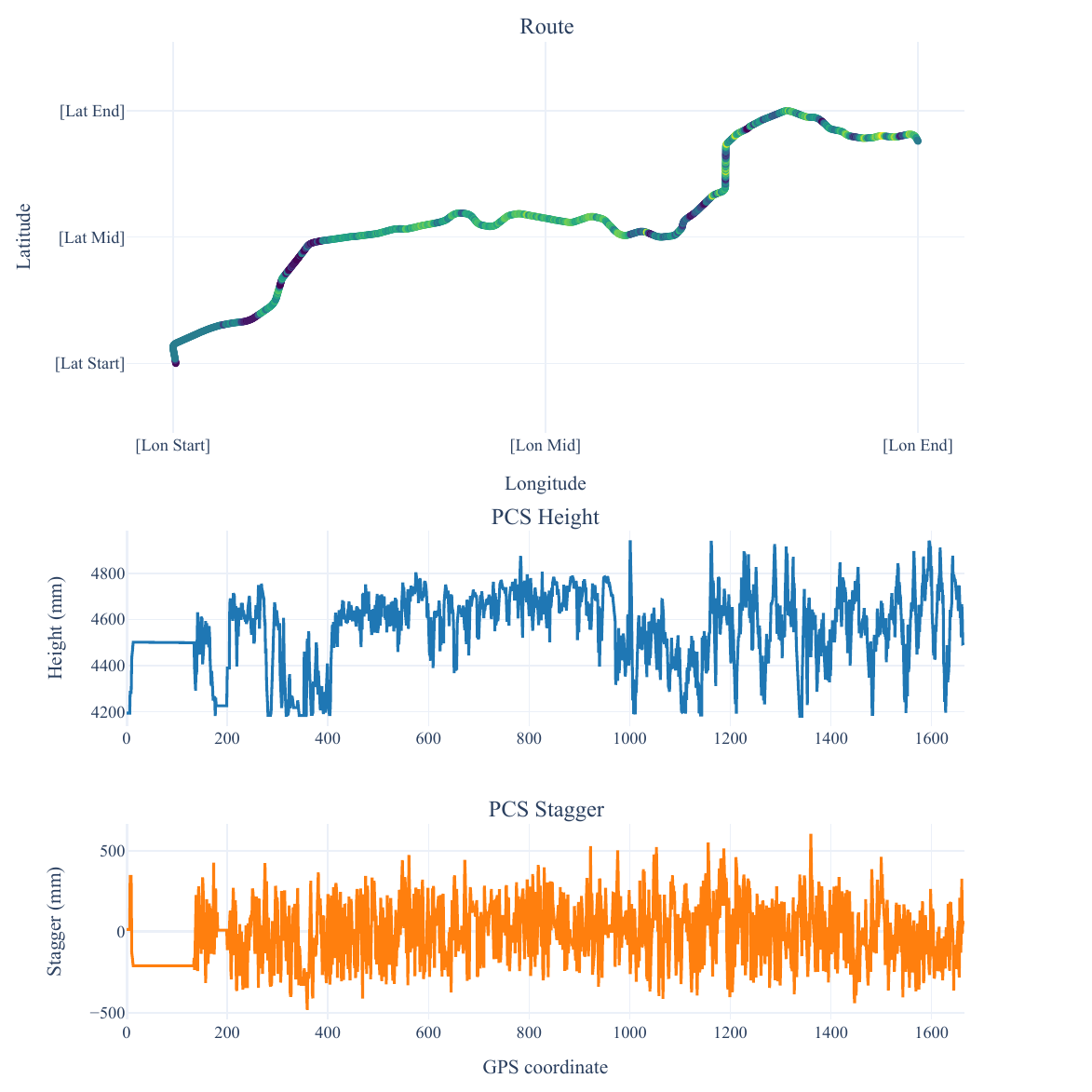}
\caption{The PCS height/stagger monitored from a train run and standardized against a reference route.}
\label{TRAIN_RUN}
\end{figure}

For \textbf{standardization} of the train runs, the raw telemetric data is spatially aligned onto a discretized master grid. First, a reference trajectory is extracted from the route's geometry to establish the baseline reference route GPS coordinates. This trajectory is then resampled using a fixed GPS coordinates interspacing of 22.0 meters --- a distance specifically derived from the onboard GPS sampling rate of 1 Hz while the train travels at an average speed of 80 km/h. The raw input data undergoes preprocessing to impute missing values, filter out stationary measurement windows, and aggregate redundant geographic readings. Finally, spatial alignment is executed by constructing a Ball Tree data structure over the raw trajectory coordinates; utilizing the Haversine distance metric, the algorithm snaps the raw GPS readings to their nearest discrete corresponding points on the nominal grid, guaranteeing uniform spatial alignment across all retained journeys. Fig.~\ref{TRAIN_RUN} shows the PCS height/stagger of a train run standardized against a reference route managed by an Italian railway company. Notably, the standardization process could not find any PCS height/stagger values aligned with the route start.

Following spatial standardization, the standardized train runs covering the continuous route undergo \textbf{segmentation}. Governed by a specific slicing parameter $L$ (which we set to 4), the overall trajectory is partitioned into discrete geographical blocks, organizing the telemetry into distinct grouped subsets ranging from segment 1 slices down to segment $n$ slices (where $n$ is equal to 4). To prepare for robust anomaly detector training, these segment slices are further divided into disjoint training, validation, and test sets. Because empirical defect data is inherently scarce, synthetic anomalies are injected exclusively into the test slices over fixed structural windows of $N = 30$ sequential points. These faults mathematically simulate two distinct infrastructure failures: a cantilever shift, which introduces a constant lateral offset to the stagger profile, and an auto-tensioning device failure, which simulates wire sag by subtracting a smooth sinusoidal penalty from the vertical height profile. Specifically, let \( N \) denote the defined defect length (where \( N=30 \)), \( I \) represent the injected intensity magnitude, and \( start \) be the randomly initialized starting index of the anomaly. The localized mathematical transformations over the defect window \( k \in [0, N-1] \) are formally defined as follows. For the cantilever shift, the original stagger sequence \( S \) is perturbed by a uniform scalar addition:
\[
S'_{start+k} = S_{start+k} + I
\]
In our case, we set this scalar addition $I$ to 750mm. For the auto-tensioning device failure, the vertical height sequence \( H \) is modified by subtracting a half-period sine wave to generate a smooth, localized sag curvature:
\[
H'_{start+k} = H_{start+k} - I \sin\left(\pi \frac{k}{N-1}\right)
\]
In this formulation, the term \( \frac{k}{N-1} \) linearly normalizes the discrete steps of the defect window to the interval \( [0, 1] \). Multiplying by \( \pi \) maps the sequence to \( [0, \pi] \) radians. This bounds the penalty within a single half-sine period, ensuring that the simulated sag begins and ends precisely at zero amplitude for a seamless geometric transition into and out of the healthy wire profile.

To perform anomaly detection on the segmented telemetry, the framework employs an LSTM-Autoencoder (LSTM-AE) architecture, uniquely designed to capture the spatial-sequential dependencies of the railway infrastructure. Because normal catenary geometry varies greatly depending on the geographical location, a distinct, localized LSTM-AE model is trained exclusively from scratch for every single defined route segment (zone) across each dataset permutation. The architecture is explicitly configured as a bidirectional LSTM network, allowing it to process the sequence of spatial coordinates both forwards and backwards, capturing the physical context on both sides of a given measurement point. Specifically, the model parameters are set as follows:
\begin{itemize}
    \item Input Features: 2 (Height and Stagger).
    \item Sequence Length: 10 sequential measurement points.
    \item Hidden Dimension: 8, expanding to a bidirectional bottleneck of 16 ($8 \times 2$).
    \item Layers: 3 stacked LSTM layers in both the encoder and decoder, utilizing a dropout rate of 0.2 to prevent overfitting on the healthy baseline data.
\end{itemize}

During the \textbf{anomaly detector training} phase, the model is fed exclusively healthy, non-anomalous data from the train set over 500 epochs using the Adam optimizer (learning rate = $1e^{-3}$) and a batch size of 32, minimizing the MSE loss. Once trained, the model reconstructs input sequences. Because it has only learned the local, healthy baseline geometry of that specific zone, it will struggle to reconstruct the injected defects present in the test set, yielding a high reconstruction error. The threshold for classifying an anomaly is dynamically established using the 99th percentile of the MSE from the separate Validation set. Any reconstructed sequence in the Test set with an MSE exceeding this localized threshold is flagged as an anomaly. The framework then exhaustively tracks standard classification metrics (AUC, F1-Score, Precision, Recall, and Confusion Matrix values) for every individual segment to aggregate the overall collective detection performance.

\begin{table}[t!]
\centering
\caption{Results of the anomaly detector with $L=4$.}
\label{tab:segmented_metrics}
\resizebox{\columnwidth}{!}{%
\begin{tabular}{lrrrrrrrr}
\toprule
Zone & AUC & F1 & Precision & Recall & TP & TN & FP & FN \\
\midrule
Zone 0 & 0.898 & 0.791 & 0.953 & 0.675 & 102 & 658 & 5 & 49 \\
Zone 1 & 0.968 & 0.921 & 0.946 & 0.897 & 140 & 650 & 8 & 16 \\
Zone 2 & 0.868 & 0.722 & 0.821 & 0.644 & 96 & 644 & 21 & 53 \\
Zone 3 & 0.967 & 0.871 & 0.954 & 0.801 & 125 & 656 & 6 & 31 \\
\bottomrule
\end{tabular}
}
\end{table}

The localized LSTM-AE architecture demonstrates robust collective anomaly detection capabilities, though performance variations exist across different geographical segments. As detailed in Table \ref{tab:segmented_metrics}, the model achieves good discrimination in Zone 1 and Zone 3, recording Area Under the Curve (AUC) scores of 0.968 and 0.967, respectively. Across all evaluated zones, the framework maintains high precision ($>0.820$), highlighting its capacity to effectively minimize false positive alarms ($FP \le 21$). However, sensitivity fluctuates depending on the segment's underlying topological baseline; Zone 2 experiences a notable reduction in recall ($0.644$), resulting in a higher incidence of false negatives ($FN = 53$). Overall, the results validate that training discrete, location-specific autoencoders successfully isolates structural catenary defects while accommodating the route's natural geometric variance.
\section{Conclusions}
\label{sec:conclusions}
In this paper, we proposed a novel framework for anomaly detection and localization within the PCS. While existing data-driven monitoring solutions effectively evaluate contact wire geometry, they fundamentally lack spatial awareness, often evaluating telemetry in a geographic vacuum. To address this critical blind spot, our approach successfully integrates spatial context by standardizing raw GPS and telemetric data onto a discretised geographical grid. By partitioning the continuous railway route into discrete segments, we established localized geometric baselines that accurately reflect the natural, healthy variations of the infrastructure at specific geographic coordinates.

Our experimental validation, leveraging real-world data from an Italian railway track, demonstrates the efficacy of localized LSTM-AE in isolating structural defects. By evaluating the framework against synthetically injected anomalies, the localized models exhibited robust collective anomaly detection capabilities, achieving high AUC and precision scores across various route segments. This suggests that training discrete, location-specific models minimizes false positive alarms caused by natural topological variance.

Ultimately, this spatially aware framework provides a highly practical tool for railway predictive maintenance, enabling operators to not only detect PCS degradation but to pinpoint its exact physical location. The proposed approach aligns with recent efforts to enhance the resilience and operational safety of railway systems through intelligent data-driven methodologies \cite{ge2025simulation,ge2025traffic}. Future research will focus on expanding the framework to process telemetry in real-time via edge computing, evaluating its robustness against a wider empirical dataset of naturally occurring faults, and exploring advanced spatial-temporal architectures, such as Graph Neural Networks, to further optimize segment boundary transitions.



\bibliographystyle{IEEEtran}
\bibliography{bibliography}

\end{document}